\pdfoutput=1

\documentclass[11pt]{article}

\usepackage[final]{EACL2023}

\usepackage{times}
\usepackage{latexsym}
\usepackage{booktabs}
\usepackage[T1]{fontenc}
\usepackage{multirow}
\usepackage{graphicx}
\usepackage{subcaption}

\usepackage[utf8]{inputenc}

\usepackage{graphicx}
\usepackage{inconsolata}
\usepackage{amsmath}
\usepackage{tablefootnote}
\usepackage{gb4e}
\noautomath

\newcommand\decreasespace{\vspace{-5pt}}

\title{Can BERT eat RuCoLA? Topological Data Analysis to Explain}

\author{
    ~\textbf{Irina Proskurina\textsuperscript{1}}, 
    ~\textbf{Irina Piontkovskaya\textsuperscript{2}},
    ~\textbf{Ekaterina Artemova\textsuperscript{3}} \\
    \textsuperscript{1}Universit{\'e} de Lyon, Lyon 2, ERIC UR 3083, France\\ 
    \textsuperscript{2}Huawei Noah’s Ark lab \\
    \textsuperscript{3}Center for Information and Language Processing (CIS), LMU Munich, Germany \\
    \textbf{Correspondence:} \href{mailto:Irina.Proskurina@univ-lyon2.fr}{Irina.Proskurina@univ-lyon2.fr}
}

\begin{document}
\maketitle
\begin{abstract}
This paper investigates how Transformer language models (LMs) fine-tuned for acceptability classification capture linguistic features. Our approach uses the best practices of topological data analysis (TDA) in NLP: we construct directed attention graphs from attention matrices, derive topological features from them, and feed them to linear classifiers. We introduce two novel features, chordality, and the matching number, and show that TDA-based classifiers outperform fine-tuning baselines. We experiment with two datasets, \textsc{CoLA} and \textsc{RuCoLA},\footnote{Arugula or rocket salad in English} in English and Russian, typologically different languages. 

On top of that, we propose several black-box introspection techniques aimed at detecting changes in the attention mode of the LMs during fine-tuning, defining the LM's prediction confidences, and associating individual heads with fine-grained grammar phenomena. 

Our results contribute to understanding the behavior of monolingual LMs in the acceptability classification task, provide insights into the functional roles of attention heads, and highlight the advantages of TDA-based approaches for analyzing LMs.
We release the code and the experimental results for further uptake.\footnote{\href{https://github.com/upunaprosk/la-tda}{https://github.com/upunaprosk/la-tda}}

\end{abstract}
\section{Introduction}
Language modelling with Transformer \cite{vaswani2017attention} has become a standard approach to acceptability judgements, providing results on par with the human baseline \cite{warstadt-etal-2019-neural}. 
The pre-trained encoders and BERT, in particular, were proven to have an advantage over other models, especially when judging the acceptability of sentences with long-distance dependencies \cite{Warstadt2019LinguisticAO}.  
Research examining linguistic knowledge of BERT-based language models (LMs) revealed that: (1) individual attention heads can store syntax, semantics or both kinds of linguistic information \cite{jo-myaeng-2020-roles,clark-etal-2019-bert}, (2) vertical, diagonal and block attention patterns could frequently repeat across the layers \cite{kovaleva-etal-2019-revealing}, and (3) fine-tuning affects the linguistic features encoding tending to lose some of the pre-trained model knowledge~\cite{miaschi-etal-2020-linguistic}. 
However, less attention has been paid to examining the grammatical knowledge of LMs in languages other than English. 
The existing work devoted to the cross-lingual probing showed that grammatical knowledge of Transformer LMs is adapted to the downstream language; in the case of Russian, the interpretation of results cannot be easily explained \cite{ravishankar-etal-2019-multilingual}. However, LMs are more insensitive towards granular perturbations when processing texts in languages with free word order, such as Russian \cite{taktasheva-etal-2021-shaking}.

In this paper, we probe the linguistic features captured by the Transformer LMs, fine-tuned for acceptability classification in Russian. 
Following recent advances in acceptability classification, we use the Russian corpus of linguistic acceptability (\textsc{RuCoLA})~\cite{mikhailov-etal-2022-rucola}, covering tense and word order violations, errors in the construction of subordinate clauses and indefinite pronoun usage, and other related grammatical phenomena.  
We provide an example of an unacceptable sentence from \textsc{RuCoLA} with a morphological violation in the pronoun usage: a possessive reflexive pronoun `svoj' (oneself's/own) instead of the 3rd person pronoun.
\begin{exe}
\ex \label{ex:intro_ex} * Eto byl pervyj chempionat mira v \textbf{svoej} kar'ere.
(``It was the first world championship in \textbf{own} career.'')
\end{exe}
Following the recently proposed Topological Data Analysis (TDA) based approach to the linguistic acceptability (LA) task \cite{cherniavskii-etal-2022-acceptability}, we construct directed attention graphs from attention matrices and then refer to the characteristics of the graphs as to the linguistic features learnt by the model. 
We extend the existing research on the acceptability classification task to the Russian language and show the advantages of the TDA-based approach to the task. 
Our main contributions are the following:~\emph{(i)}~we investigate the monolingual behaviour of LMs in acceptability classification tasks in the Russian and English languages, using a TDA-based approach, 
~\emph{(ii)}~we introduce new topological features and outperform previously established baselines,
~\emph{(iii)}~we suggest a new TDA-based approach for measuring the distance between pre-trained and fine-tuned LMs with large and base configurations. \emph{(iv)}~We determine the roles of attention heads in the context of LA tasks in Russian and English. 

Our initial hypothesis is that there is a difference in the structure of attention graphs between the languages, especially for the sentences with morphological, syntactic, and semantic violations. 
We analyze the relationship between models by comparing the features of the attention graphs.
To the best of our knowledge, our research is one of the first attempts to analyse the differences in monolingual LMs fine-tuned on acceptability classification corpora in Russian and English, using the TDA-based approach.



\section{Related Work}
\paragraph{Acceptability Classification.} First studies performed acceptability classification with statistical machine learning methods, rule-based systems, and context-free grammars~\cite{cherry-quirk-2008-discriminative,wagner2009judging,post-2011-judging}. Alternative approaches use threshold scoring functions to estimate the likelihood of a sentence \cite{lau-etal-2020-furiously}. Recent research has been centered on the ability of omnipresent Transformer LMs to judge acceptability \cite{wang-etal-2018-glue}, to probe for their grammar acquisition \cite{zhang-etal-2021-need}, and evaluate semantic correctness in language generation \cite{batra-etal-2021-building}. 
In this project, we develop acceptability classification methods and apply them to datasets in two different languages, English and Russian.

\paragraph{Topological Data Analysis (TDA) in NLP.}  Recent work uses TDA to explore the inner workings of LMs.  \citet{kushnareva-etal-2021-artificial} derive TDA features from attention maps to build artificial text detection. \citet{colombo-etal-2021-automatic} introduce \textsc{BaryScore}, an automatic evaluation metric for text generation that relies on Wasserstein distance and barycenters. \citet{chauhan2022bertops} develop a scoring function which captures the homology of the high-dimensional hidden representations, and is aimed at test accuracy prediction. 
We extend the set of persistent features proposed by \citet{cherniavskii-etal-2022-acceptability} for acceptability classification and conduct an extensive analysis of how the persistent features contribute to the classifier's performance.

\paragraph{How do LMs change via fine-tuning?} There have been two streams of studies of how fine-tuning affects the inner working of LM's: (i) what do sub-word representation capture and (ii) what are the functional roles of attention heads? The experimental techniques include similarity analysis between the weights of source and fine-tuned checkpoints \cite{clark-etal-2019-bert}, training probing classifiers \cite{durrani-etal-2021-transfer}, computing feature importance scores \cite{atanasova-etal-2020-diagnostic}, the dimensionality reduction of sub-word representations \cite{alammar-2021-ecco}. Findings help to improve fine-tuning procedures by modifying loss functions  \cite{elazar-etal-2021-measuring} and provide techniques for explaining LMs' predictions \cite{danilevsky-etal-2020-survey}. Our approach reveals the linguistic competence of attention heads by associating head-specific persistent features with fine-grained linguistic phenomena. 

\section{Methodology}
We follow \citealp{warstadt-etal-2019-neural} and treat the LA task as a supervised classification problem.
We fine-tune Transformer LMs to approximate the function that maps an input sentence to a target class: acceptable or unacceptable. 
\subsection{Extracted Features}\label{sec:tda_method}
Given an input text, we extract output attention matrices from Transformer LMs and follow \citealp{kushnareva-etal-2021-artificial} to compute three types of persistent features over them.

\begin{figure*}[h!]
\centering
\begin{subfigure}{0.3\textwidth}
    \includegraphics[width=0.8\textwidth]{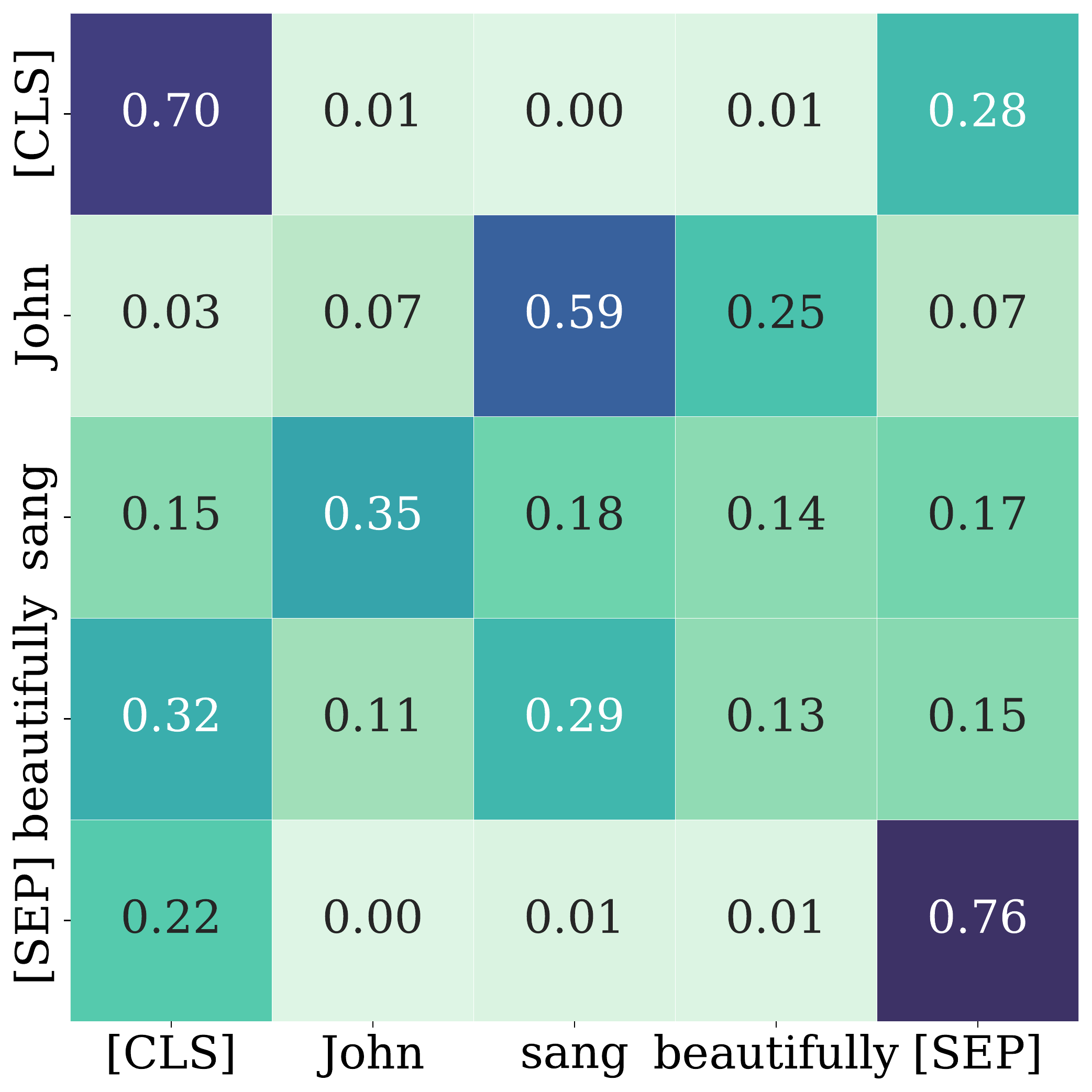}
    \caption{}
    \label{fig:first}
\end{subfigure}
\hfill
\begin{subfigure}{0.3\textwidth}
    \includegraphics[width=1.1\textwidth]{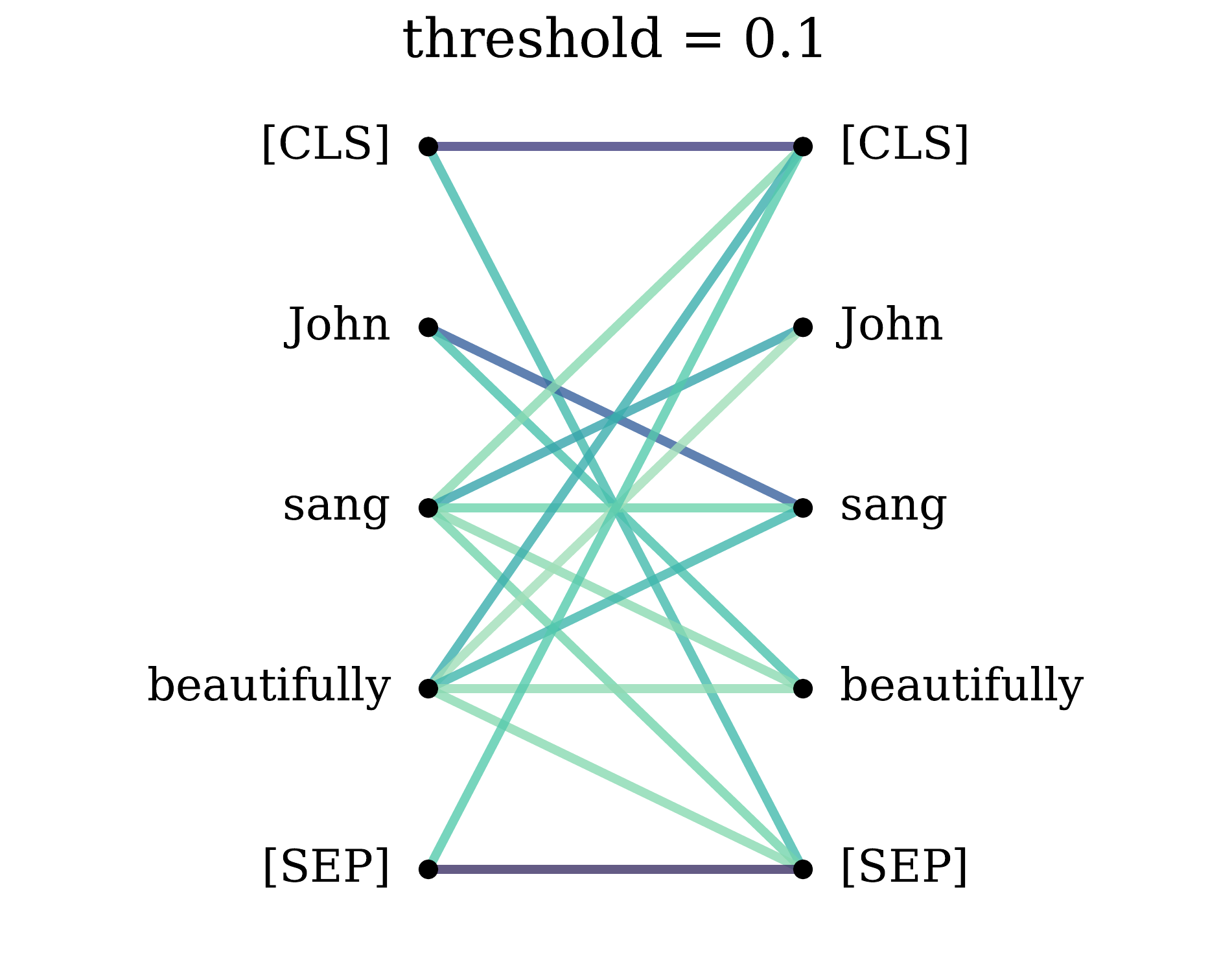}
    \caption{}
    \label{fig:second}
\end{subfigure}
\hfill
\begin{subfigure}{0.36\textwidth}
    \includegraphics[width=1.1\textwidth]{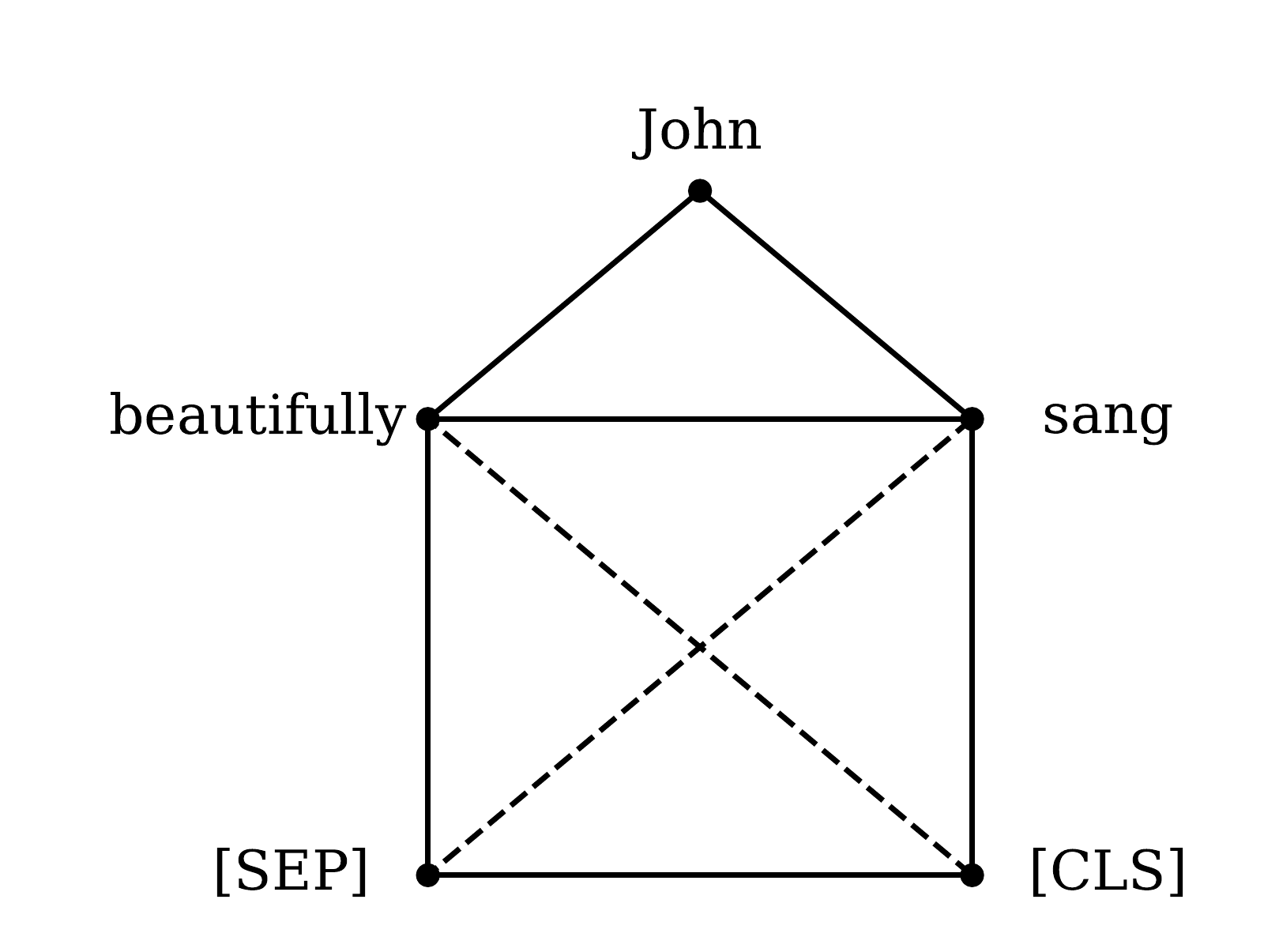}
    \caption{}
    \label{fig:third}
\end{subfigure}

\caption{An example of an attention map (a) and the corresponding bipartite (b) and attention (c) graphs for the \textsc{CoLA} sentence \textit{``John sang beautifully''}.  The graphs are constructed with a threshold equal to 0.1.}
\label{fig:ex_features}
\end{figure*}



\textbf{Topological} features are properties of attention graphs.
We provide an example of an attention graph constructed upon the attention matrix in \autoref{fig:ex_features}.
An adjacency matrix of attention graph $A=(a_{ij})_{n\times n}$ is obtained from the attention matrix $W=(w_{ij})_{n\times n}$, using a pre-defined threshold $thr$: 
\begin{equation*}
a_{ij} = 
 \begin{cases}
   1 &\text{if $w_{ij}\geq thr$}\\
   0 &\text{otherwise,}
 \end{cases}
\end{equation*}
where $w_{ij}$ is an attention weight between tokens $i$ and $j$ and $n$ is the number of tokens in the input sequence. Each token corresponds to a graph node.
Features of directed attention graphs include the number of strongly connected components, edges, simple cycles and average vertex degree. 
The properties of undirected graphs include the first two Betti numbers: the number of connected components and the number of simple cycles. 
We propose two new features of the undirected attention graphs: the matching number and the chordality. 
The matching number is the maximum matching size in the graph, i.e. the largest possible set of edges with no common nodes.

Consider an attention matrix depicted in \autoref{fig:first} and a simple undirected attention graph (\autoref{fig:third}) constructed based on the bipartite graph (\autoref{fig:second}) with a threshold of 0.1.
The matching number of that attention graph is equal to two.
One example of a maximum matching in that graph is a set of edges: \{(\textit{John} - \textit{sang}), (\textsc{[SEP]} - \textsc{[CLS]})\}. 
That matching is maximum because there are no more edges that are not incident to the already matched 4 nodes (tokens).
The chordality is a binary feature showing whether the attention graph is chordal; that is, whether the attention graph does not contain induced cycles of a length greater than~3.
For example, the plotted graph in \autoref{fig:third} is chordal because it does not contain induced cycles with more than 3 edges. 
If there were no dotted edges (chords) in the graph, there would be a cycle \textsc{[SEP]}-\textit{beautifully}-\textit{sang}-\textsc{[CLS]}-\textsc{[SEP]} of length 4, meaning that the graph is not chordal. 

We expect these novel features to express syntax phenomena of the input text. 
The chordality feature could carry information about subject-verb-object triplets. The maximum matching can correspond to matching sentence segments (subordinate clauses, adverbials, participles, introductory phrases, etc.). 

\textbf{Features derived from barcodes} include descriptive characteristics of $0/1$-dimensional barcodes and reflect the survival (death and birth) of connected components and edges throughout the filtration. 

\textbf{Distance-to-pattern} features measure the distance between attention matrices and identity matrices of pre-defined attention patterns, such as attention to the first token \textsc{[CLS]} and to the last \textsc{[SEP]} of the sequence, attention to previous and next token and to punctuation marks \cite{clark-etal-2019-bert}.
We use a publicly available implementation to compute features.\footnote {\href{https://github.com/danchern97/tda4atd}{https://github.com/danchern97/tda4atd}}

\subsection{Experimental Framework}
\paragraph{Data}
We use two publicly available LA benchmarks in two typologically different languages: Russian (\textsc{RuCoLA}; \citealp{mikhailov-etal-2022-rucola}) and English (\textsc{CoLA}; \citealp{warstadt-etal-2019-neural}).
Both selected corpora consist of in- and out-of-domain data and contain sentences collected from linguistics publications; each is marked as acceptable or unacceptable. 
Unacceptable sentences are annotated with syntactic, morphological and semantic phenomena violated in them.
\textsc{RuCoLA}, in addition, covers synthetically generated data by generative LMs.
We provide examples of acceptable sentences from observed corpora (\ref{ex:cola_acc}, \ref{ex:rucola_acc}) along with sentences with semantic violations (\ref{ex:cola_unacc}, \ref{ex:rucola_unacc}).
\begin{exe}
\ex \begin{xlist}
    \ex\label{ex:cola_acc} The dog bit the cat.
    \ex\label{ex:cola_unacc} * The \textbf{soundly and furry} cat slept.
    \end{xlist}
\ex \begin{xlist}
    \ex\label{ex:rucola_acc} Koshki byli svyashchennymi zhivotnymi v Drevnem Egipte. 
     (``Cats were sacred animals in ancient Egypt.'')
    \ex\label{ex:rucola_unacc} * \textbf{Bliz} kresla na nebol'shom kovrike lezhala koshka.
    (``\textbf{Outside of} an armchair on a small rug a cat was lying.'')
    \end{xlist}
\end{exe}
\decreasespace
\autoref{tab:corpora-summary} (\autoref{sec:app_data}) reports statistics of the used corpora.
For per-category evaluation, we use \textsc{RuCoLA} error annotations, and for \textsc{CoLA}, we use minor grammatical phenomena annotations to group erroneous sentences. 
We provide more details in \autoref{tab:en-cola-groups} (\autoref{sec:app_data}). 

\paragraph{Models}
Our baseline model architectures, fine-tuning and evaluation scripts are taken from the Transformers library \cite{wolf-etal-2020-transformers}. 
We use the following case-sensitive monolingual Transformer LMs for the experiments: 
(1) base size En-BERT\footnote{\href{https://huggingface.co/bert-base-cased}{hf.co/bert-base-cased}}~\cite{devlin-etal-2019-bert} and Ru-BERT,\footnote{\href{https://huggingface.co/sberbank-ai/ruBert-base}{hf.co/sberbank-ai/ruBert-base}} (2) large size En-RoBERTa\footnote{\href{https://huggingface.co/roberta-large}{hf.co/roberta-large}}~\cite{liu2019roberta} and Ru-RoBERTa.\footnote{\href{https://huggingface.co/sberbank-ai/ruRoberta-large}{hf.co/sberbank-ai/ruRoberta-Large}}
To estimate the effect of fine-tuning, we compare two types of models: pre-trained LMs with frozen weights (frozen) and fine-tuned LMs on the training sets.
Transformer LMs are fine-tuned for 5 epochs on in-domain training data, with a batch size of 32 and an optimal set of hyper-parameters determined by the authors of the datasets.
To mitigate class imbalance, we use weighted cross-entropy loss.
We provide fine-tuning details in \autoref{tab:bert_hyperparameters} (\autoref{sec:app_data}). 
\paragraph{TDA Classifiers}
We extract a range of persistent (TDA) features listed in Section~\ref{sec:tda_method} from Transformer LMs and refer to them as training features fed to a linear classifier.  
We reduce the feature space dimensionality with principal component analysis (PCA).
Next, we train Logistic Regression classifiers with adjusted class weights on the reduced feature space.
We iterate over a range of inverse regularization parameter values $C\in \{10^{-3},10^{-2}, 0.1\}$ and the number of principal components $\#PC\in [10,20\dots200]$. 
We choose the value 200 as the upper bound of the PC grid to ensure that the number of latent features is at least two times less than the size of the  in-domain development (IDD) or out-of-domain development (OODD) sets. 
We tune hyper-parameters to maximize the classifier performance on the IDD set.
We compare the performance of two feature sets, by reporting results of classifiers trained on (i) basic TDA features by \citealp{kushnareva-etal-2021-artificial} (dubbed as TDA) and  (ii) TDA features with two novel features added (dubbed as TDA$_{ext})$. 
\subsection{Evaluation}
\label{sec:method_evaluation}
\paragraph{Performance Metrics}
Following \citealp{warstadt-etal-2019-neural}, we measure performance with Accuracy (Acc.) and Matthews Correlation Coefficient (MCC).
MCC is used as the main performance metric for finding hyperparameters, evaluating trained models, and adjusting the decision threshold.
\paragraph{Fine-tuning Effect} 

We estimate changes in attention weights between pre-trained and fine-tuned LMs with two methods. 
First, we follow \citealp{hao-etal-2020-investigating} and employ Jensen-Shannon (JS) divergence:
\begin{equation*}
\begin{aligned}
D_{JS}(M_t||M_0)&=  \frac{1}{N}\frac{1}{H}\sum_{n=1}^{N}\sum_{h=1}^{H}\frac{1}{W}\sum_{i=1}^{K}\\
& D_{JS}(W_t^h(token_i)||W_0^h(token_i))
\end{aligned}
\end{equation*}
where $M_t$ and $M_0$ are fine-tuned and frozen models respectively, $N$ is number of sentences,
$H$ is a number of attention heads ($H=12$ for base-configuration LMs,  $H=24$ for large LMs),
$K$ is the number of tokens in the sentence $n$, and $W_t^h(token_i)$ is an attention weight of attention head $h$ at token $i$ in model $M_t$.

Second, we estimate the difference between attention graphs as an average correlation distance between the TDA$_{ext}$ features across attention heads:
\begin{equation*}
\begin{aligned}
D_{TDA}(M_t,M_0)= \frac{1}{H}\sum_{h=1}^{H}\frac{1}{F}\sum_{f=1}^{F}D_{corr}(V_{tf}^h,V_{0f}^h)
\end{aligned}
\end{equation*}
where $F$ is the number of features, $V_{tf}^h$ are values of the  feature $f$, computed over attention matrix $W_t^h$, extracted from the model $M_t$.
\section{Results}

\begin{table}[t!]
\scriptsize
\centering
\newcommand{\hsp}{\hspace{4pt}}
\setlength{\tabcolsep}{2pt}
\begin{tabular}{@{}lcccc@{}} 
\toprule
\multirow{3}{*}{\textbf{Model}}  &
\multicolumn{2}{c}{\textbf{Fine-tuned LMs}} & \multicolumn{2}{c}{\textbf{Frozen LMs}} \\[0.4ex]

&  \textbf{IDD} & \textbf{OODD}  & \textbf{IDD} & \textbf{OODD} \\ [0.4ex]

&  \textbf{Acc.} \hsp \textbf{MCC} & \textbf{Acc.} \hsp \textbf{MCC}  & \textbf{Acc.} \hsp \textbf{MCC} & \textbf{Acc.} \hsp \textbf{MCC} \\
\midrule

\multicolumn{5}{c}{\textbf{\textsc{RuCoLA}}} \\ 
\midrule
Ru-BERT  & 80.3 \hsp 0.420 & 75.1 \hsp 0.438  &  62.4 \hsp 0.079 & 54.7 \hsp 0.112 \\
+ TDA & 80.1 \hsp 0.440  &  75.1 \hsp 0.447 & 76.5 \hsp 0.314 & 62.3 \hsp 0.253 \\ 
+ TDA$_{ext}$ & 80.1 \hsp 0.478 & 73.2 \hsp 0.440 & 76.7 \hsp 0.331 & 62.6 \hsp 0.270 \\ 
\midrule
Ru-RoBERTa & 83.5 \hsp 0.530 & 79.3 \hsp 0.530  &  72.8 \hsp 0.313 & 58.1 \hsp 0.241 \\
+ TDA & \underline{85.0} \hsp \underline{0.581}  & \textbf{81.0} \hsp \textbf{0.584} & \underline{77.0} \hsp \underline{0.374} & \textbf{64.7} \hsp \underline{0.343} \\ 
+ TDA$_{ext}$ & \textbf{85.7} \hsp \textbf{0.594} & \underline{80.1} \hsp \underline{0.558} & \textbf{77.2} \hsp \textbf{0.391} & \underline{64.2} \hsp \textbf{0.358} \\ 
\midrule
\multicolumn{5}{c}{\textbf{\textsc{CoLA}}} \\
\midrule
En-BERT & 85.0 \hsp 0.634 & 82.0 \hsp 0.561  & 62.6 \hsp 0.039 & 64.3 \hsp 0.124 \\
+ TDA  & 85.6 \hsp 0.649 & 81.4 \hsp 0.548 & 77.0 \hsp 0.484 & 68.4 \hsp 0.335 \\ 
+ TDA$_{ext}$ & \textbf{88.2} \hsp \textbf{0.726} & 81.0 \hsp 0.556 & \underline{81.4} \hsp 0.543 & 73.1 \hsp 0.369 \\ 
\midrule
En-RoBERTa & 87.3 \hsp 0.692 & \textbf{84.9} \hsp \textbf{0.637}  & 74.0 \hsp 0.317 & 75.0 \hsp 0.362 \\
+ TDA  & 86.3 \hsp 0.680 & \underline{83.5} \hsp \underline{0.620} & 81.2 \hsp \underline{0.543} & \textbf{78.5} \hsp \underline{0.464} \\ 
+ TDA$_{ext}$  & \underline{87.3} \hsp \underline{0.695} & 83.1 \hsp 0.604 &  \textbf{83.1} \hsp \textbf{0.604} & \underline{77.3} \hsp \textbf{0.476} \\ 
\bottomrule
\end{tabular}
\caption{Acceptability classification results of monolingual LMs and linear classifiers trained on the sets of features by the benchmark. \textbf{IDD}=in domain development set. \textbf{OODD}=out of domain development set. TDA$_{ext}$=TDA features+chordality and the matching number.
The best score is in bold, and the second-best one is underlined.
}
\label{tab:cola_results}
\end{table}

\begin{table*}
\centering
\begin{tabular}{lccccc}
\toprule
\textbf{Model} & \textbf{Acceptable} & \textbf{Hallucination} & \textbf{Morphology} & \textbf{Semantics}& \textbf{Syntax}\\
\midrule
Ru-BERT & 92.1 & 53.9 & 20.0 & 25.0 & 55.7\\
+TDA$_{ext}$ & 80.6 & 73.9 & 53.9 & 46.6 & 76.6\\
\midrule
En-BERT & 94.3 & 68.5 & 69.4 & 63.0 & 55.6\\
+TDA$_{ext}$ & 84.5 & 78.8 & 82.5 & 76.3 & 73.0\\
\bottomrule
\end{tabular}
\caption{\label{tab:scores_cat_overall}
Overall per-category recall by the benchmark.
}
\end{table*}
\subsection{Acceptability Classification}
\autoref{tab:cola_results} reports LA classification results.
Linear classifiers trained on the TDA features boost Transformer LMs performance; that trend is consistent across all models, with the MCC score gain of +0.252 at most for the Russian LMs and a more substantial +0.504 increase falling on En-BERT. Proposed chordality and matching number features are beneficial and help improve performance, proving that they capture linguistic information. 


Unlike base LMs, large frozen LMs exhibit grammatical knowledge even  before fine-tuning. Base LMs' MCC scores fluctuate around zero, while large LMs achieve at least 0.3 MCC. 

That observation aligns with the recent works showing  that pre-trained large En-RoBERTa can achieve competitive scores without further fine-tuning in tasks such as lexical complexity prediction \cite{rao-etal-2021-rg}.

At the same time, TDA classifiers outperform fine-tuned models by a minor margin enhancing scores by 
at best +0.064 MCC for Russian and +0.092 MCC for English. 
We believe that fine-tuning may cause the LM to lose general grammatical skills and forget language phenomena that are not present in the fine-tuning set \cite{miaschi-etal-2020-linguistic}.
Thus, the features extracted from the fine-tuned models may require a thorough feature selection with non-linear models to mitigate feature redundancy issues.
TDA classifies for \textsc{RuCoLA} achieve scores on par with the baseline LMs. However,  for \textsc{CoLA}, the  TDA$_{ext}$ classifier coupled with En-RoBERTa outperforms the baseline.
We report classification results on OOD test data in \autoref{tab:rucola_test} and \autoref{tab:cola_test}, Appendix \ref{sec:app_res}.

\subsection{Sensitivity to Violation Categories}
\label{sec:per_cat_sensitivity}


Next, we analyze gains in recall by TDA classifiers with respect to violation category. 
\autoref{tab:scores_cat_overall} reports scores of Ru-BERT and En-BERT baselines and TDA classifiers averaged between IDD and OODD sets with respect to 5  grammatical violations.
TDA classifiers outperform LMs in unacceptable sentences; that uptrend holds for both languages, while there is a drop for acceptable sentences.

In contrast to English, the TDA$_{ext}$ classifier trained on Ru-BERT features is more sensitive to syntactic violations reaching the overall 76.6 recall; that is, the increase in the score is around 20 recall points, compared to fine-tuned Ru-BERT. 
As for the rest grammar categories, the TDA$_{ext}$  classifier outperforms the fine-tuned Ru-BERT by a large margin, especially in sentences with word-level morphological violations, where the recall of Ru-BERT is more than doubled.

Next, we manually analyze the errors of the fine-tuned Ru-BERT and our classifier TDA$_{ext}$ in OODD sentences in Russian.
First, we compare the unacceptable sentences, which are  misclassified by Ru-BERT but correctly classified by the TDA$_{ext}$ classifier. 
We find  that the error span in OODD sentences is relatively short, with at most three tokens.
In particular, in these sentences, such violations as non-existing words are most often encountered, the misuse of which is quite common among native speakers
(\ref{ex:rucola_morph}, word formation error `ekhaj'), local inverse word order (\ref{ex:rucola_synt}), or nonsense (\ref{ex:rucola_sem}). 
Common false predictions of both models include long sentences that mix grammatical phenomena, contain long-distance agreement violations and complex errors in punctuation.

\begin{exe}
\ex \begin{xlist}
    \ex\label{ex:rucola_morph} * A ty \textbf{ekhaj} pryamo k direktoru teatrov.
(``\textbf{You should gotta to} the director of theatres.'')
    \ex\label{ex:rucola_synt} * V etom lesu \textbf{vodyatsya volki}. \\
    (``There are \textbf{in this forest wolves}.'')
    \ex\label{ex:rucola_sem} * Oni chitali moi zhaloby \textbf{na sebya}. (``They read my complaints \textbf{onto themselves}.'')
    \end{xlist}
\end{exe}

\begin{figure*}[t!]
    \centering
    \includegraphics[width=0.9\linewidth]{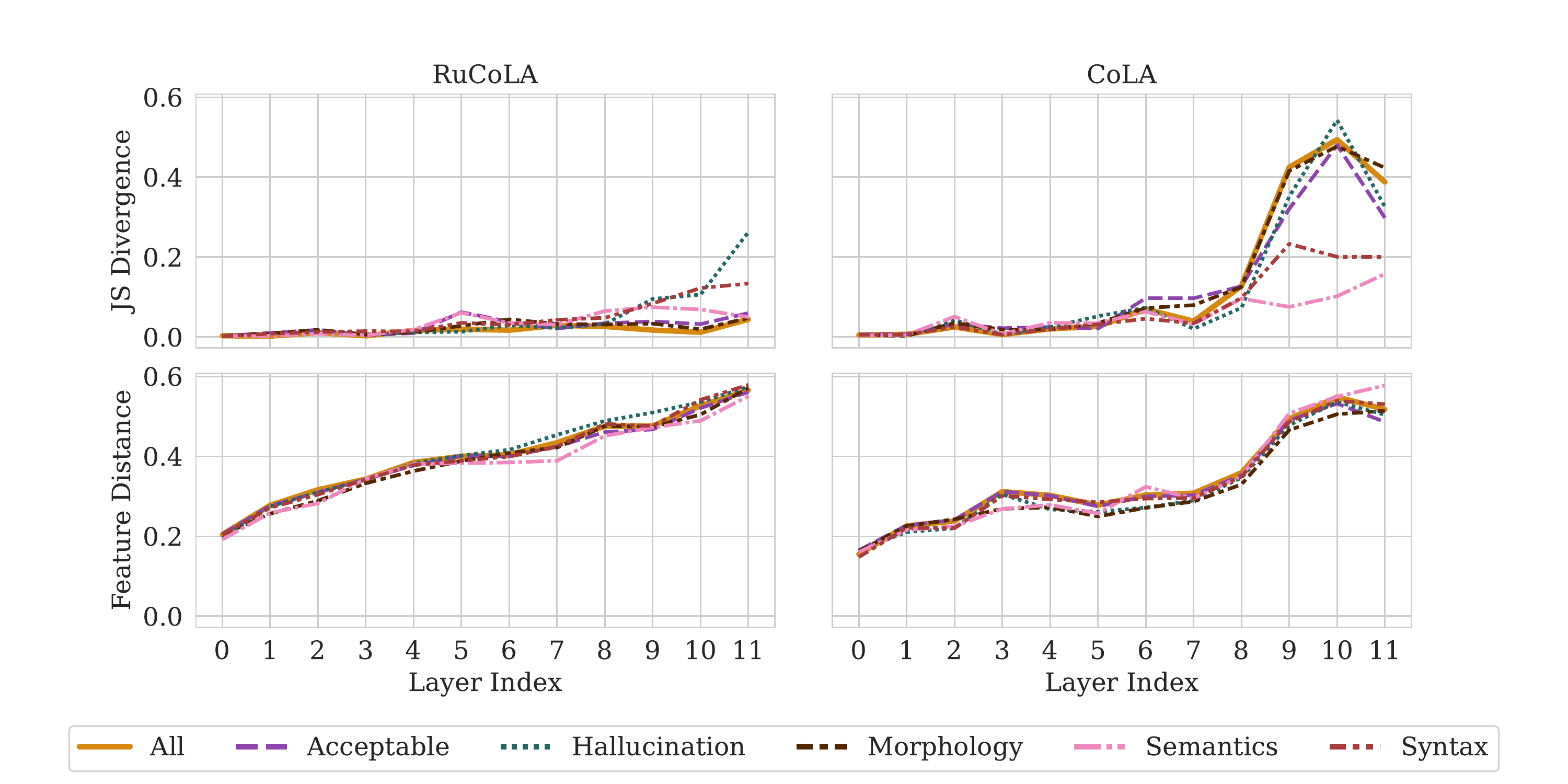}
    \caption{Per-layer feature distance and JS divergence of attention scores between the frozen and fine-tuned Ru-BERT and En-BERT.}
    \label{fig:bert_dist}
\end{figure*}
The domain shift from ID to OOD introduces new types of unacceptable phenomena are not present in ID data. Overall, the scores for OOD data are lower than for ID data  (\autoref{tab:scores_cat_overall}, \autoref{tab:scores_cat_dev}, Appendix \ref{sec:app_res}). Hence  LMs do not generalize well to unseen unacceptable phenomena and have little knowledge about the unseen linguistic properties.

\subsection{Fine-tuning Effect}

We investigate the dynamics of LM fine-tuning and measure per layer distance between TDA$_{ext}$ features extracted from frozen and fine-tuned LMs on OODD subsets (\S\ref{sec:method_evaluation}). 
\autoref{fig:bert_dist} illustrates layer-wise feature distance and JS divergence for Ru-BERT and En-BERT (\autoref{fig:roberta_dist}, Appendix \ref{subsec:fteff} for large models).
Overall, we find that the distance between features rises steadily from the bottom to higher layers, whilst for English LMs, the most noticeable changes occur only in the last four layers. 
That observation implies that there is a noticeable difference in fine-tuning dynamics between En-BERT and Ru-BERT. 

For both languages, the feature distance trend differs from JS divergence, especially in the first six layers. 
This indicates that the TDA$_{ext}$ features can be used to detect minor changes in the lower layers that are poorly expressed when using the JS divergence.
For example, TDA-based distance is sensitive to small changes in the attention weights at lower predefined thresholds where large attention weights remain unchanged. JS divergence is not capable of capturing such cases.

The distance between features is uniform with respect to the violation category.
The trends for acceptable and unacceptable sentences almost coincide, albeit there are noticeable differences in JS divergence. 
For Russian models, JS divergence in sentences with syntactic violations and hallucinations is more evident in higher layers compared to other categories. 
In turn, the JS divergence for English shows that the attention mode is more consistent with the frozen En-BERT on the sentences with semantic and syntactic violations; for acceptable and other sentences, the peak is reached at the penultimate layer. 
Similar to LMs with the base configuration, there is a steady increase in feature dissimilarity across all the layers, while for English, the main changes appear in higher layers.

\subsection{Head Importance}
\begin{table*}[ht]
\centering
\begin{center}
\begin{tabular}{lcccc}
\toprule
\textbf{Error type} & \textbf{Sentence} & \textbf{Feature} & \textbf{Head}\\
    \midrule
    \multirow{2}{*}{Morphology}
    & Recept \textbf{chipy} s syrom, \textbf{maniokom} i yajcami.
    &\multirow{2}{*}{$c_{thr0.25}$} 
    & \multirow{2}{*}{(9,5)} \\
    &  (``Recipe of \textbf{cheps} with cheese, \textbf{maniokom} and eggs.'') & &\\
    \midrule
    \multirow{2}{*}{Syntax}
    & Bylo nachato \textbf{stroit' novyj rajon}.
    & \multirow{2}{*}{$c_{thr0.1}$}
    & \multirow{2}{*}{(9,5)}\\
    &  (``\textbf{Of new district building} was started.'')
    & &\\
    \midrule
    \multirow{2}{*}{Semantics}
    & Vchera v dva chasa magazin \textbf{zakryt}. 
    & \multirow{2}{*}{\textsc{[CLS]}}
    & \multirow{2}{*}{(11,0)}\\
    & (``The store \textbf{closing} at two o'clock yesterday.'')
    & &\\
\bottomrule
\end{tabular}
\end{center}
\caption{\label{tab:feature_imp_error} Examples of the most important Ru-BERT TDA$_{ext}$ features for judging \textsc{RuCoLA} unacceptable sentences by error type. $c$ = the number of simple cycles in a graph, $thr$ = threshold used for constructing attention graph, \textsc{[CLS]} = distance-to-\textsc{[CLS]}-token.}
\end{table*}

We probe linguistic phenomena with the help of persistent features: we exploit the learnt feature weights in the linear classifiers (Appendix~\ref{sec:feat_imp}). The higher the weight of the feature, the more it contributes to the final prediction. We aggregate features derived from each head: the importance of the head is derived as a number of important features.
We define two types of heads: (1) heads that  contribute the most to true positive and true negative predictions (i.e. correct predictions), dubbed as agreeing heads, and (2)  heads that contribute the most to false negative and false positive predictions (i.e. classifier's errors), dubbed as disagreeing heads. 
First, we explore the importance of each individual head. \autoref{fig:attn_heads}, Appendix~\ref{subsec:head} shows how important the head is for the final prediction. 
En-BERT and Ru-BERT have similar patterns for the heads of type (1) as the most useful features for Ru-BERT are housed in middle to higher layers. 
For En-BERT, these tend to be localized mostly in the last two layers. 

Next, we compute the feature importance with respect to the violation category.  
Heads of middle layers contribute more to detecting syntactic and morphology violations in English and Russian.
Heads of type (2) do not overlap with the heads of type (1) with a few exceptions, which are head 10 and head 0 from the last layer of Ru-BERT and En-BERT, respectively.
Judging by the number of type (2) heads Ru-BERT struggles the most to distinguish sentences with hallucinations from acceptable sentences. This might be due to multiple reasons: (i) hallucinated sentences are not seen during training, (ii) hallucinated sentences are mainly well-formed but semantically incorrect, so there are no surface or syntactical clues to rely on.

Next, we determine the set of sentences that are the most challenging for the TDA classifier and, thus, the corresponding LM since TDA features are extracted from its attention map. 
To do so, we define the LM's confidence as the sum of absolute feature weights for predicting acceptable and unacceptable classes. The lower the score, the more confused the LM is and the more attention heads tend to disagree with the desired prediction.  
We consider those sentences challenging that obtain the lowest confidence scores. 
The most challenging sentences are long, consist of multiple clauses and contain terms or named entities, see the unacceptable sentence in \ref{ex:rucola_att_heads} for example.
For the sake of completeness, we conduct the same analysis for \textsc{CoLA} sentences and provide an example of the most confusing sentence for TDA$_{ext}$ classifier (\ref{ex:cola_att_heads}). The results align well. The most challenging sentences contain long-distance dependencies and named entities.
\begin{exe}
\ex \label{ex:rucola_att_heads} * Eta gruppa obnaruzhila \textbf{(nepravil'no) chto severnyj predel} Merrimak byl bliz togo, chto teper' izvestno kak ozero \textbf{Vinnipesuki} v \textbf{N'yu-Gempshire}.\\ 
(``This group found \textbf{(poorly), that the northern watershed} of the Merrimack was near what is now known as Lake \textbf{Vinnipesaukee} in \textbf{New Gampshire}.'')
\ex\label{ex:cola_att_heads} * Gould's performance of Bach on the piano doesn't please me anywhere as much as \textbf{Ross's on the harpsichord}.

\end{exe}

Finally,  we explore the feature contribution on the sentence level.
Our TDA-based approach allows explaining predictions for every single sentence. 
To this end, the contribution (=importance) of each feature is the feature value multiplied by the learnt weight of the linear classifier. 
We observe the following patterns across unacceptable sentences in Russian and Ru-BERT:
\begin{enumerate}
    \item Distance-to-pattern features appear to be useful for classifying unacceptable sentences with word-level violations, including spelling, punctuation, and agreement errors;
    \item Topological features and features derived from barcodes contribute equally to more complicated grammatical phenomena.
\end{enumerate}

\autoref{tab:feature_imp_error} provides examples of unacceptable sentences along with the feature importance values.
Chordality, the matching number, the number of simple cycles, and the average vertex degree derived at thresholds 0.1 or 0.25 frequently become important to predict unacceptable sentences in Russian. 
Similarly, the average number of vertex degrees has the most discriminative power for English and En-BERT.
Important features are housed across different layers in the LMs.
For English, the most important features are extracted from the last layer, while for Russian, they appear at the earliest at layer 6.

However, when it comes to the discrepancy in attention graphs between acceptable and unacceptable sentences, we find the following common for both languages. 
The number of connected components in attention graphs for unacceptable sentences is larger at the lowest and the highest thresholds. 
At the highest threshold, these components consist of one token; at the lowest one, they consist of a few ones. 
It means that the values of attention maps in unacceptable sentences do not deviate much from each other. 
On the contrary, for acceptable sentences, there is a tendency to put the most attention weight on a single token, which is usually the sentence's head verb.
In terms of the TDA feature values, this effect can be seen as the sign of the correlation coefficient between the feature value and the target class correlation. 
Thus, there is an obvious shift towards positive correlation at a threshold of 0.5 for average vertex degree features (\autoref{fig:feature_thr}).

To sum up, such an analysis helps better explain the classifiers' prediction. Since persistent features are attributed to individual heads, we can trace the role and importance of each head. A fine-grained annotation of language phenomena allows us to associate specific linguistic skills with individual heads.


\section{Conclusion}

In this paper, we adopt and improve methods for acceptability classification by using best practices from topological data analysis (TDA). 
We showcase the developed methods in two typologically different languages by using the datasets in English and Russian, \textsc{CoLA} and \textsc{RuCoLa}, respectively. 
In particular, we introduce two novel features, chordality and the matching number, and compare the performance of TDA-based classifiers to fine-tuning.
TDA-based classifiers boost the performance of pre-trained language models.


TDA-based classifiers have advantages over LM fine-tuning because they are more interpretable and help to introspect the inner workings of LMs. 
To this end, we introduce a TDA feature-based distance measure to detect changes in the attention mode of LMs during fine-tuning. 
This distance measure is sensitive even to small changes occurring at the bottom layers of LMs that are not detected by the widespread Jensen-Shannon divergence. 
What is more important, we show how TDA features reveal the functional roles of attention heads. 
We compare heads that contribute to making correct and incorrect predictions based on their importance. 
This way we discover heads that store information about word order, word derivation, and complex semantic phenomena in unacceptable sentences and heads that attend to acceptable sentences.  

Given the sentence, we evaluate the prediction confidence based on the contribution of the features.  
We determine the set of sentences in which LMs are less confident and find that those sentences usually consist of multiple clauses and frequently include named entities.   
Finally, we find a distinct pattern  that is frequently present in the attention maps of unacceptable sentences in English and Russian.

We hope that our results shed light on the performance of LMs in Russian and English and help understanding their fine-tuning dynamics and the functional roles of attention heads. We are excited to see the adoption of TDA by NLP practitioners to other languages and downstream problems. 


%

\section*{Limitations} 

\paragraph{Acceptability judgments datasets} Acceptability judgments datasets use linguistic literature as source of unacceptable sentences. Such approach is subject to criticism on two counts: (i) 
 the reliability and reproducibility of acceptability judgments  \cite{gibson2013need,culicover2010quantitative,sprouse2013empirical,linzen2018reliability},  (ii) representativeness, as linguists' judgments may not reflect the errors that speakers tend to produce~\cite{dkabrowska2010naive}.

\paragraph{Computational complexity}
The computation complexity of the proposed features is linear. 
For chordality features, we rely on the implementation of linear $O(|E|+|V|)$ time algorithm \cite{tarjan1984simple}, where $|E|$ and $|V|$ are the numbers of edges and nodes, respectively.
We use a greedy algorithm with linear complexity $O(|E|)$ to find the maximum matching.
When calculating simple cycles with the exponential-time algorithm (in the worst case), we use a constraint equal to 500 to do an early stopping.
We suggest that simple cycles features are less informative when that value is exceeded.
\citealp{kushnareva-etal-2021-artificial} discuss the time complexity of the rest features.

\section*{Acknowledgements}
We thank Laida Kushanareva,  Daniil Cherniavskii, Vladislav Mikhailov, Serguei Barannikov, Alexander Bernstein, and Dmitri Piontkovski for their comments at its early stages, and we thank Max Ryabinin for providing scripts to process the \textsc{RuCoLA} dataset. 

\bibliography{anthology,custom}
\bibliographystyle{acl_natbib}

\appendix

\newpage 
\clearpage

\onecolumn 

\section{Experiment Setup}
\label{sec:app_data}
\begin{table}[th!]
\small
\centering
\newcommand{\hsp}{\hspace{4pt}}
\begin{tabular}{lrr} 
\toprule
 &  \textbf{\textsc{CoLA}} & \textbf{\textsc{RuCoLA}}  \\
\midrule
Language & English & Russian \\
Data type & Real & Real, Synthetic\\
$\alpha$ & 0.86  & 0.89\\
\# Train sent. & 8551 & 7869 \\
\# Dev sent. & 527   & 983 \\
\# Test sent. &  516 & 1804 \\
\% & 70.5 & 71.8\\
\bottomrule
\end{tabular}
\caption{Statistics of language acceptability corpora. $\alpha$ = Average annotator agreement rate. \% = Percentage of acceptable sentences. }
\label{tab:corpora-summary}
\end{table}

\begin{table}[th!]
\centering
\small 
\newcommand{\hsp}{\hspace{4pt}}
\begin{tabular}{p{3cm}|p{2cm}} 
\toprule
\textbf{Grammatical feature} & {\textbf{Error type}}   \\
\midrule
Extra/Missing Word  & Hallucination \\
Semantic Violations &  Semantics \\
Infl/Agr Violations  &  Morphology\\
Other  &  Syntax\\
\bottomrule
\end{tabular}
\caption{\textsc{CoLA} features aggregated by error type. Infl/Agr =Inflection and
Agreement. Other=the rest of grammar violation phenomena present in \textsc{CoLA} annotation for unacceptable sentences.}
\label{tab:en-cola-groups}
\end{table}

\begin{table}[th!]
\centering
\small 
\begin{tabular}{lccc} 
\toprule
\textbf{Model} & \textbf{Learning rate}  &  \textbf{Weight decay} \\
\midrule
En-BERT & $3\cdot 10^{-5}$ & $0.01$  \\
En-RoBERTa & $2\cdot 10^{-5}$ & $10^{-4}$\\ 
Ru-BERT & $3\cdot 10^{-5}$ & $0.1$\\
Ru-RoBERTa &  $10^{-5}$ & $10^{-4}$\\
\bottomrule
\end{tabular}
\caption{Hyperparameter values used for finetuning transformers.}
\label{tab:bert_hyperparameters}
\end{table}

\newpage
\clearpage

\section{Experiment Results}

\subsection{ Linguistic Acceptability  Classification}\label{sec:app_res}
\begin{table}[th!]
\small
\centering
\newcommand{\hsp}{\hspace{4pt}}
\begin{tabular}{@{}lcc@{}} 
\toprule
\multirow{2}{*}{\textbf{Model}} &  \textbf{Expert} & \textbf{Machine} \\ [0.4ex]
&  \textbf{Acc.} \hsp \textbf{MCC} & \textbf{Acc.} \hsp \textbf{MCC}\\

\midrule
Ru-BERT & 77 \hsp 0.37 & 75 \hsp 0.44 \\ 
+ TDA$_{ext}$  & 75 \hsp 0.39 & 72 \hsp 0.42 \\
\midrule
Ru-RoBERTa  & 84 \hsp 0.55 & 80 \hsp 0.56  \\ 
+ TDA$_{ext}$  & 83 \hsp 0.53 & 80 \hsp 0.56  \\ 
\bottomrule
\end{tabular}
\caption{Linguistic acceptability  classification results of monolingual LMs and linear classifiers on \textsc{RuCoLA} out of domain test set by source.\tablefootnote{\href{https://rucola-benchmark.com}{https://rucola-benchmark.com}}}
\label{tab:rucola_test}
\end{table}
\begin{table}[th!]
\small
\centering
\begin{tabular}{@{}lc@{}} 
\toprule
\textbf{Model} &  \textbf{MCC}\\
\midrule
En-BERT & 0.509\\
+ TDA$_{ext}$  & 0.469\\
\midrule
En-RoBERTa  & 0.608\\
+ TDA$_{ext}$  & 0.616\\
\bottomrule
\end{tabular}
\caption{Acceptability classification results of monolingual LMs and linear classifiers on \textsc{CoLA} out of domain test set \cite{cola-out-of-domain-open-evaluation}.}
\label{tab:cola_test}

\end{table}

\begin{table*}[th!]
\centering
\small
\begin{tabular}{lccccc}
\midrule
\textbf{Model} & \textbf{Acceptable} & \textbf{Hallucination} & \textbf{Morphology} & \textbf{Semantics}& \textbf{Syntax}\\
\midrule
\multicolumn{6}{c}{\textbf{\textsc{RuCoLA}} \textbf{IDD}} \\ 
\midrule
Ru-BERT & 93.9 & - & 12.5 & 24.0 & 56.0\\
+TDA$_{ext}$ & 86.2 & - & 56.2 & 45.0 & 75.4 \\
\midrule
Ru-RoBERTa & 95.9 & - & 50.0 & 37.0 & 70.9\\
+TDA$_{ext}$ & 
96.3 & - & 31.2 & 34.0 & 72.4 \\
\midrule
\multicolumn{6}{c}{\textbf{\textsc{RuCoLA}} \textbf{OODD}} \\ 
\midrule
Ru-BERT & 90.3 & 53.9 & 26.6 & 25.9 & 55.4\\
+TDA$_{ext}$ & 
75.0 & 73.9 & 51.6 & 48.1 & 77.7 \\
\midrule
Ru-RoBERTa & 90.9 & 64.3 & 54.7 & 42.0 & 75.5\\
+TDA$_{ext}$ & 89.9 & 63.9 & 53.1 & 39.5 & 71.4 \\
\midrule
\multicolumn{6}{c}{\textbf{\textsc{CoLA}} \textbf{IDD}} \\ 
\midrule
En-BERT & 94.8 & 65.0 & 69.0 & 72.2 & 61.2\\
+TDA$_{ext}$ & 87.9 & 77.5 & 86.2 & 83.3 & 82.4 \\
\midrule
En-RoBERTa & 94.8 & 72.5 & 88.9 & 75.9 & 64.7 \\
+TDA$_{ext}$ & 87.3 & 75.0 & 79.3 & 88.9 & 70.6 \\
\midrule
\multicolumn{6}{c}{\textbf{\textsc{CoLA}} \textbf{OODD}} \\ 
\midrule
En-BERT & 93.8 & 72.0 & 69.7 & 53.8 & 50.0 \\
+TDA$_{ext}$  & 81.0 & 80.0 & 78.8 & 69.2 & 63.5\\
\midrule
En-RoBERTa & 93.5 & 76.0 & 87.9 & 76.9 & 56.2 \\
+TDA$_{ext}$ & 83.1 & 80.0 & 81.8 & 92.3 & 63.5 \\
\bottomrule
\end{tabular}
\caption{\label{tab:scores_cat_dev}
Per-category recall on the IDD and OODD sets by benchmark.
}
\end{table*}

\newpage
\clearpage

\subsection{Fine-tuning effect} \label{subsec:fteff}
\begin{figure*}[th!]
    \centering
    \includegraphics[width=0.9\linewidth]{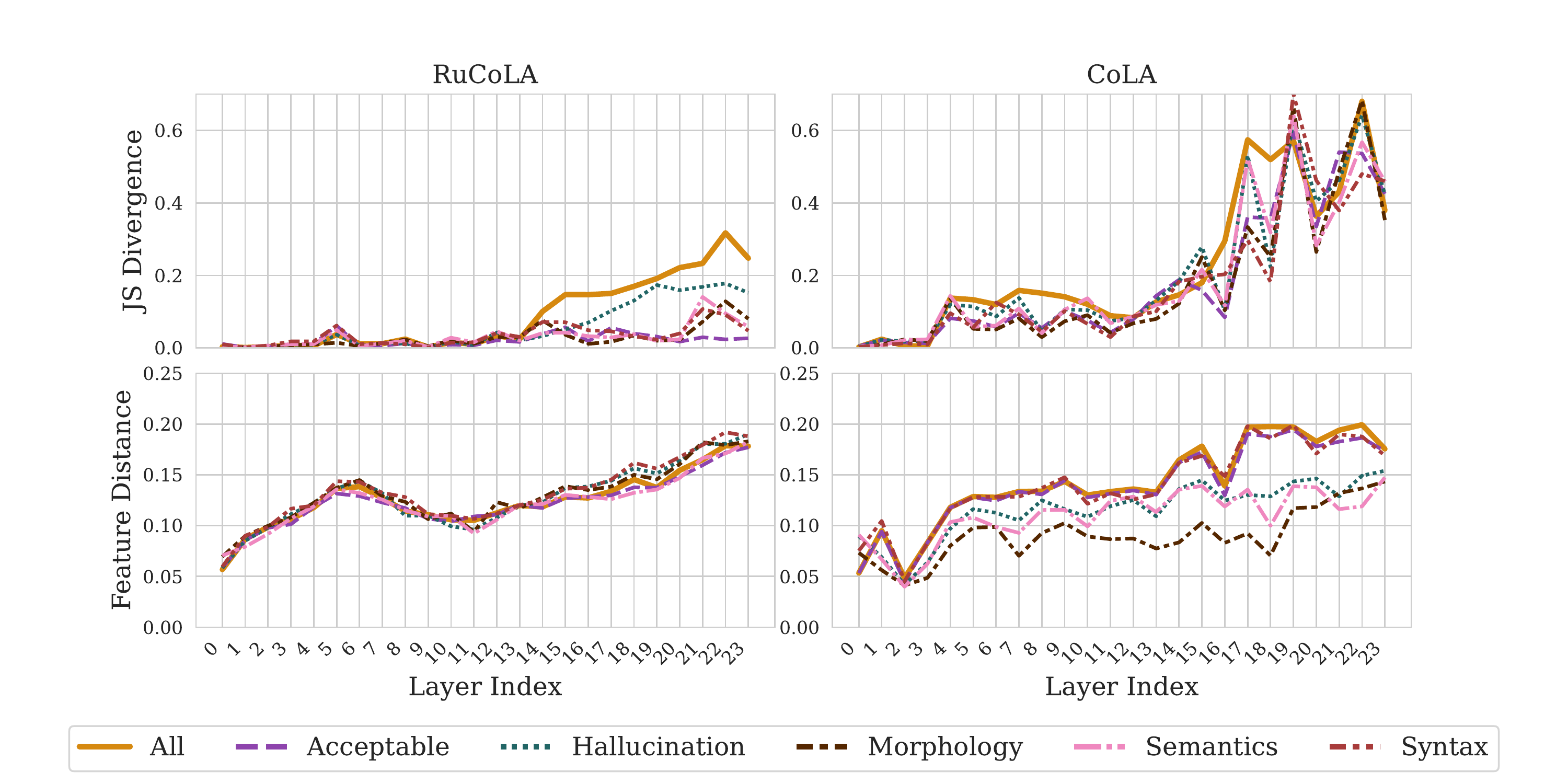}
    \caption{Per-layer feature distance and Jensen-Shannon divergence of attention scores between the frozen and fine-tuned
Ru-RoBERTa and En-RoBERTa.}
    \label{fig:roberta_dist}
\end{figure*}
\subsection{Feature Importance}
\label{sec:feat_imp}
Consider a linear classifier with L1 regularization, then the output probability for the sentence $i$ is:
$$p_i \sim \text{exp}(X_{0i}^TC^Tw + c),$$
where $X_{0 i}$ are the input TDA features, $C$ is the principal component matrix, $w^T$ is a vector of PCs coefficients in the decision function, and $c$ is the added bias.  $C^Tw$ is the feature contribution to prediction.

\begin{figure*}
\subsection{The Roles of Attention Heads} \label{subsec:head}
\centering
\includegraphics[width=0.8\textwidth]{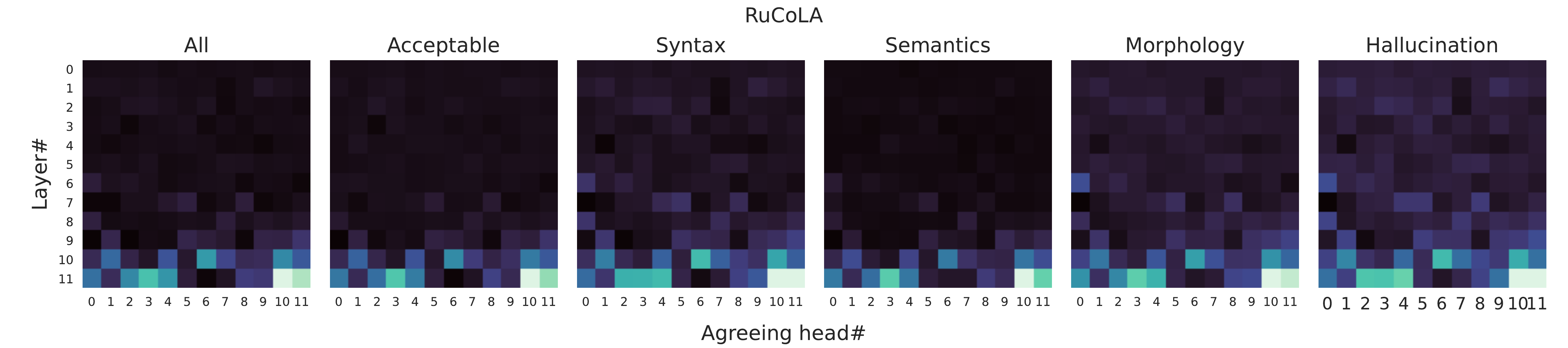}\\
\includegraphics[width=0.8\textwidth]{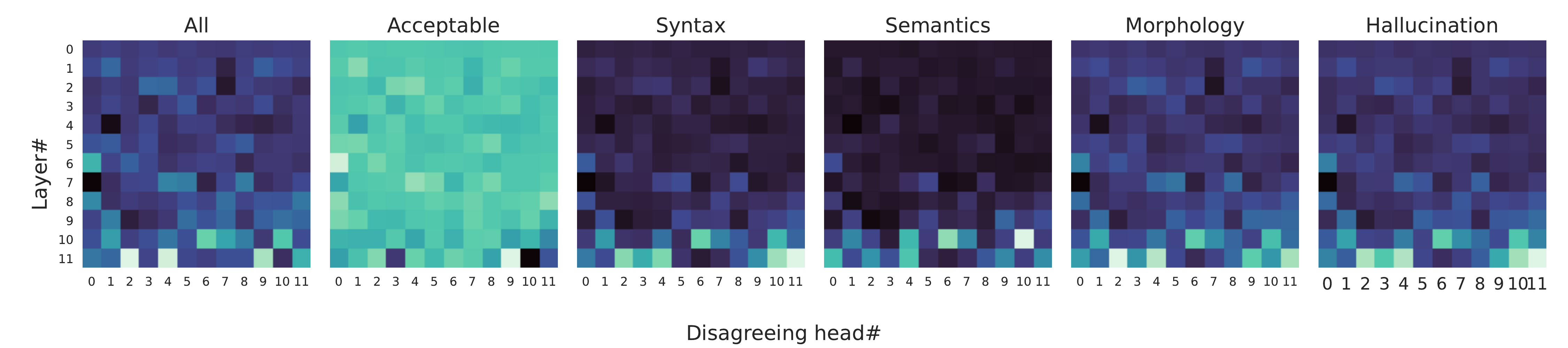}\\
\medskip
\includegraphics[width=0.8\textwidth]{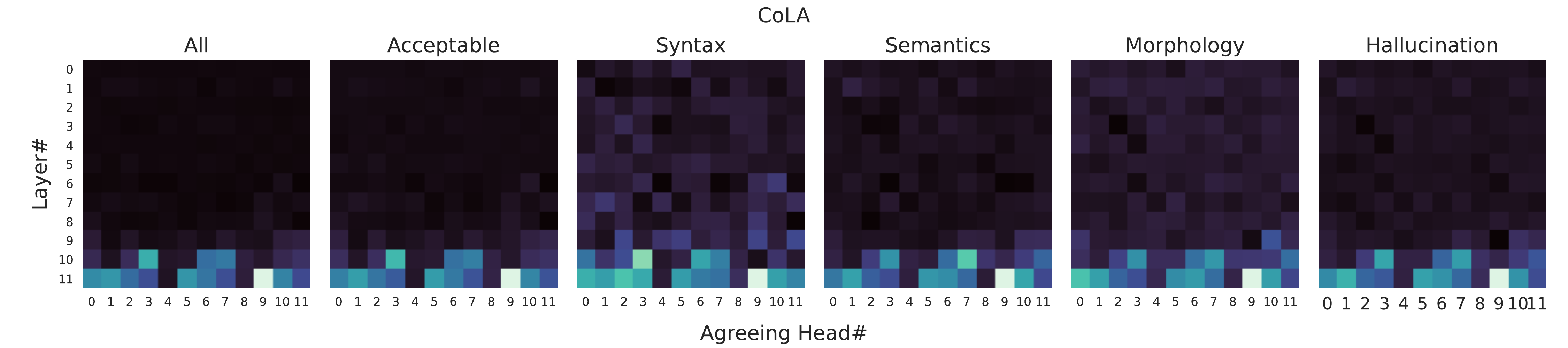}
\includegraphics[width=0.8\textwidth]{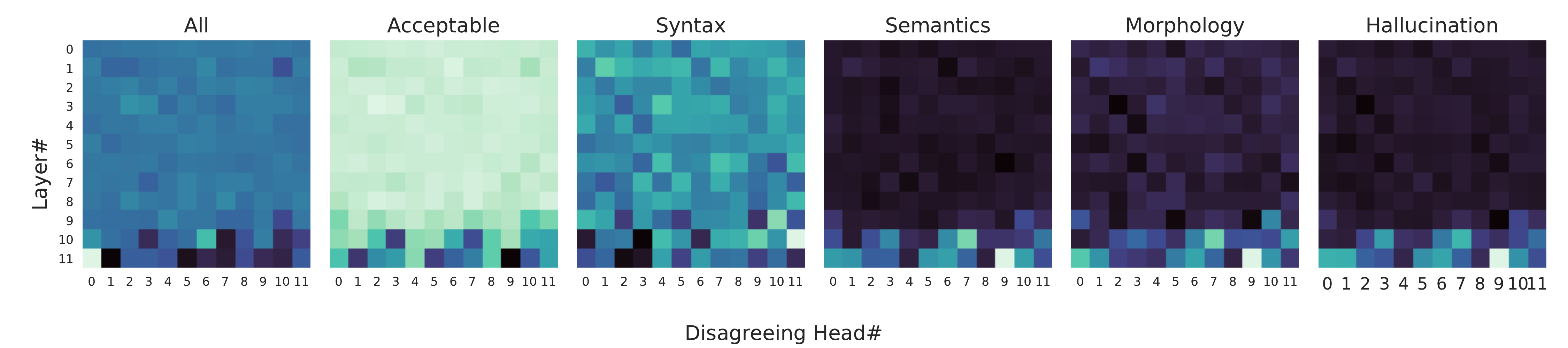}
\caption{Mean feature weights in TDA$_{ext}$ classifiers with respect to the dataset. TDA$_{ext}$ are extracted from fine-tuned Ru-BERT and En-BERT, respectively. 
Features of an \textit{agreeing head} contribute to correct prediction. Features of an \textit{disagreeing head} contribute to incorrect prediction. Brighter colors stand for higher mean feature weights.}
\label{fig:attn_heads}

\centering
\includegraphics[width=0.8\textwidth]{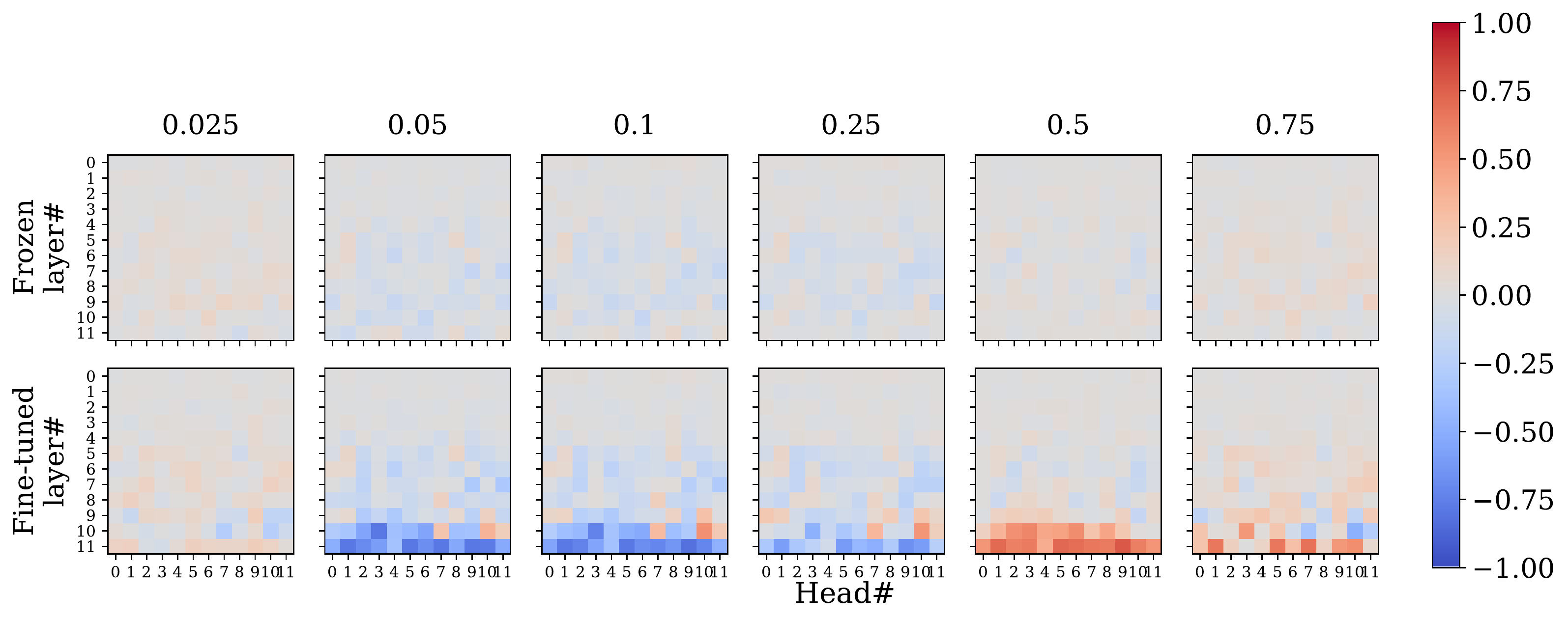}\\
\medskip
\caption{Correlation coefficients between average vertex degree features and target labels for frozen and fine-tuned Ru-BERT by the threshold used to construct attention graph.}
\label{fig:feature_thr}
\end{figure*}
\end{document}